\documentclass[preprint,3p]{elsarticle}
\usepackage{times}
\usepackage{epsfig}
\usepackage{setspace}
\usepackage{graphicx}
\usepackage{textcomp}
\usepackage{amsmath,amsthm,amssymb,amsfonts}
\usepackage{array}
\usepackage{multirow}
\usepackage{booktabs}
\usepackage{color}
\usepackage{amsfonts}
\usepackage{lineno}
\usepackage{lscape}
\usepackage{bm}
\usepackage{algpseudocode}
\usepackage{threeparttable}
\usepackage{algorithm}
\usepackage{makecell}
\usepackage{siunitx}
\usepackage{pifont}
\usepackage[figuresright]{rotating}
\usepackage[colorlinks,
            linkcolor=red,
            anchorcolor=blue,
            citecolor=green
            ]{hyperref}

\DeclareMathOperator*{\argmax}{arg\,max}

\journal{Journal of \LaTeX\ Templates}

\begin{document}

\begin{frontmatter}

  \title{Iterative pseudo-labeling based adaptive copy-paste supervision for semi-supervised tumor segmentation}

  \author[1,2]{Qiangguo Jin}
  \author[3]{Hui Cui}
  \author[1]{Junbo Wang}
  \author[4]{Changming Sun}
  \author[1]{Yimiao He}
  \author[5]{Ping Xuan}
  \author[6]{Linlin Wang}
  \author[7]{Cong Cong}
  \author[8,9]{Leyi Wei\corref{mycorrespondingauthor}}
  \author[10]{Ran Su\corref{mycorrespondingauthor}}
  \cortext[mycorrespondingauthor]{Corresponding author}

  \address[1]{School of Software, Northwestern Polytechnical University, Shaanxi, China}
  \address[2]{Yangtze River Delta Research Institute of Northwestern Polytechnical University, Taicang, China}
  \address[3]{Department of Computer Science and Information Technology, La Trobe University, Melbourne, Australia}
  \address[4]{CSIRO Data61, Sydney, Australia}
  \address[5]{Department of Computer Science, School of Engineering, Shantou University, Guangdong, China}
  \address[6]{Shandong Cancer Hospital and Institute, Shandong First Medical University and Shandong Academy of Medical Sciences, Shandong, China}
  \address[7]{Australian Institute of Health Innovation (AIHI), Macquarie University, Australia}
  \address[8]{School of Software, Shandong University, Shandong, China}
  \address[9]{AIDD, Faculty of Applied Science, Macao Polytechnic University, Macao SAR, China}
  \address[10]{School of Computer Software, College of Intelligence and Computing, Tianjin University, Tianjin, China}

\begin{abstract}
Semi-supervised learning (SSL) has attracted considerable attention in medical image processing. The latest SSL methods use a combination of consistency regularization and pseudo-labeling to achieve remarkable success. However, most existing SSL studies focus on segmenting large organs, neglecting the challenging scenarios where there are numerous tumors or tumors of small volume. Furthermore, the extensive capabilities of data augmentation strategies, particularly in the context of both labeled and unlabeled data, have yet to be thoroughly investigated. To tackle these challenges, we introduce a straightforward yet effective approach, termed iterative pseudo-labeling based adaptive copy-paste supervision (IPA-CP), for tumor segmentation in CT scans. IPA-CP incorporates a two-way uncertainty based adaptive augmentation mechanism, aiming to inject tumor uncertainties present in the mean teacher architecture into adaptive augmentation. Additionally, IPA-CP employs an iterative pseudo-label transition strategy to generate more robust and informative pseudo labels for the unlabeled samples. Extensive experiments on both in-house and public datasets show that our framework outperforms state-of-the-art SSL methods in medical image segmentation. Ablation study results demonstrate the effectiveness of our technical contributions.
\end{abstract}


\begin{keyword}
Semi-supervised learning \sep two-way uncertainty \sep iterative pseudo-labeling \sep copy-paste
\end{keyword}

\end{frontmatter}


\section{Introduction}
\label{sec:introduction}
\begin{figure}
  \centering
  \includegraphics[scale=0.27]{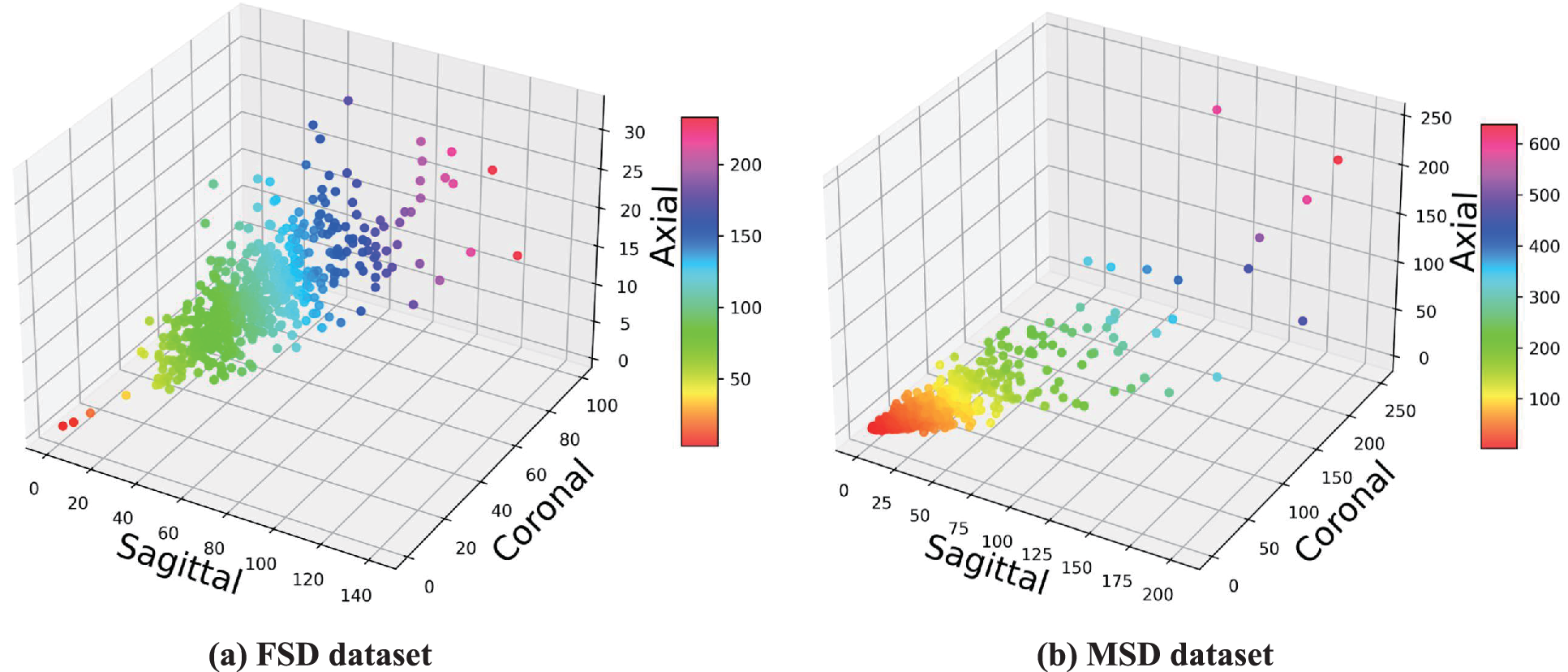}
  \caption{Comparative visualization of tumor sizes in two tumor datasets after preprocessing, encompassing sagittal, coronal, and axial perspectives. (a) The fistula esophageal tumor (FSD) dataset demonstrates a concentration of smaller tumors
  . (b) The liver tumor (MSD) dataset illustrates a broader distribution of tumor sizes.
  }
  \label{fig:overview of tumor data}
\end{figure}

Accurate tumor segmentation in medical images plays a crucial role in the diagnosis, treatment planning, and monitoring of cancer patients~\cite{li2022development}. This is particularly evident in its application to quantifying tumor growth, delineating tumor boundaries, and aiding in precise radiation therapy planning, such as in cases of esophageal and liver tumors~\cite{ye2022multi,gruber2019joint}. However, tumor segmentation is inherently challenging, primarily due to the variability in tumor size (as illustrated in Fig.~\ref{fig:overview of tumor data}), shape, appearance, and contrast with surrounding tissues. These factors frequently result in difficulties in achieving consistent and accurate delineation~\cite{patel2023multi}. In this context, automated tumor segmentation from computed tomography (CT) scans becomes an essential tool, providing the possibility of consistent, reproducible, and efficient tumor delineation.

Current medical tumor segmentation methodologies predominantly depend on extensively annotated datasets, which are essential for training accurate and reliable deep learning models. 
However, acquiring detailed annotations poses a considerable challenge, as it requires expertise from trained radiologists or oncologists. The process of annotating tumors is not only labor-intensive but also subject to variability due to the intricate nature of tumor morphology and other factors on variability including interpersonal differences, lack of standardization, and subjectivity~\cite{bai2017semi}. As a result, the scarcity and difficulty in obtaining high-quality annotated tumor data significantly impede the development and optimization of segmentation models. In this context, semi-supervised learning (SSL) has become an increasingly vital methodology in the realm of medical image segmentation. SSL leverages both labeled and unlabeled data, offering an effective solution to the challenges posed by the annotation bottleneck. 

Current SSL methods in medical image processing can be roughly divided into three categories: consistency regularization based, self-training and pseudo-labeling based, and adversarial methods. Consistency regularization focuses on ensuring stable model predictions under various input perturbations and penalizing the model to ignore irrelevant variations in the data~\cite{ouali2020semi}. Self-training and pseudo-labeling approaches iteratively refine model predictions on unlabeled data, using enhanced predictions as pseudo labels for further training, thereby gradually improving the model's accuracy and robustness~\cite{chen2021semi}. Lastly, adversarial models utilize the underlying data distribution of both labeled and unlabeled samples to improve segmentation, creating synthetic yet realistic images for training~\cite{lei2022semi}. Each of these categories contributes uniquely to the advancement of semi-supervised segmentation, offering varied solutions to the inherent challenges of limited availability with labeled data.

Despite the significant advancements in recent years, current SSL methods for medical image segmentation still face several limitations. (1) Many approaches, such as UA-MT~\cite{yu2019uncertainty}, bidirectional copy-paste (BCP)~\cite{bai2023bidirectional}, and its variants~\cite{yu2024enhancing,song2024sdcl,wang2025data}, are primarily designed for segmenting larger organs but struggle with various small-volume tumors, limiting their applicability in real-world clinical scenarios. (2) Other methods, such as MCF~\cite{wang2023mcf}, URPC~\cite{luo2022semi}, and MC-Net+~\cite{wu2021semi,wu2022mutual}, focus on integrating additional network components and hybrid loss functions, often increasing computational complexity without fully leveraging simple data augmentation strategies to improve consistency between labeled and unlabeled data. (3) Some SSL approaches~\cite{wang2022dual} train labeled and unlabeled data separately, failing to ensure consistency across both domains, which can negatively impact model generalization. (4) Pseudo-labeling strategies remain imperfect, as current refinement techniques~\cite{zhang2022dynamic,basak2023pseudo,su2024mutual} may introduce incorrect predictions due to variations in tumor morphology and the inherent discrepancy between teacher and student models. (5) The Mixup and CutMix methods, which randomly blend large image regions, may overlook uncertainties in tumor boundaries. Moreover, while BCP-based methods~\cite{bai2023bidirectional,yu2024enhancing,song2024sdcl,wang2025data} enhance data diversity by bidirectionally pasting regions between labeled and unlabeled data, they either introduce computational complexity through diffusion models or fail to dynamically adjust augmentation strength based on prediction confidence, which is particularly crucial for small tumors. These limitations highlight the need for a simple and adaptive strategy that effectively enhances small tumor segmentation, optimizes augmentation strategies, and improves pseudo-labeling robustness.

To address these challenges, we propose a straightforward yet effective approach, iterative pseudo-labeling based adaptive copy-paste supervision (IPA-CP)~\footnote{Source code is released at https://github.com/BioMedIA-repo/IPA-CP.git}, for semi-supervised tumor segmentation, as shown in Fig.~\ref{fig:overview of framework}. Without additional parameters or modules, IPA-CP utilizes two-way uncertainties, i.e., divergence in prediction from the student model relative to the teacher model and vice versa, as a measure of prediction variance within the mean teacher architecture. We inject two-way uncertainties into an adaptive augmentation strategy, highlighting prediction discrepancies between the teacher and student models. Further, we employ an iterative process, gradually aligning the student's pseudo labels with the teacher's predictions, thereby generating more robust and informative pseudo labels for unlabeled samples. Finally, our method incorporates a bidirectional copy-paste strategy to effectively utilize both labeled and unlabeled data under supervised training.
The major innovations and contributions of the proposed method include:
\begin{itemize}
  \item \textbf{Research problem to be solved:} Unlike existing SSL approaches that primarily focus on segmenting large organs, we specifically address a more pioneer problem of small tumor segmentation. This problem has been largely overlooked in prior SSL research, despite its critical importance in clinical applications.
  \item \textbf{Innovative technical design:} We propose IPA-CP, an innovative SSL method for tumor segmentation that seamlessly integrates labeled and unlabeled data, enhancing performance without requiring additional network components or extra training losses. IPA-CP introduces two novel strategies to improve SSL for small tumors: (1) a two-way uncertainty-based adaptive augmentation, which dynamically adjusts augmentation strength based on uncertainty estimates, ensuring effective contributions from both labeled and unlabeled data; and (2) an iterative pseudo-label transition strategy, which progressively refines pseudo labels to mitigate noise accumulation and enhance supervision quality for unlabeled data.
  \item \textbf{Effectiveness and generalizability:} To comprehensively evaluate IPA-CP, we construct a large in-house FSD dataset specifically for small tumor segmentation. Additionally, we validate our method's generalization capability on the publicly available MSD dataset, demonstrating its robustness across different tumor types.
\end{itemize}

\section{Related Work}
\label{sec:related_works}
\subsection{Fully supervised medical tumor segmentation}
Recent advancements in deep learning have substantially influenced medical image segmentation, particularly in tumor segmentation tasks~\cite{gruber2019joint,patel2023multi,ranjbarzadeh2021brain,jiang2022deep}. Broadly, fully supervised tumor segmentation techniques can be categorized into two groups: U-Net~\cite{ronneberger2015u} based architectures and non-U-Net based models. For instance, Wang et~al.~\cite{wang2022relax} proposed an architecture based on U-Net with ResBlock for fully automatic brain tumor segmentation in MRI images. Recently, UNetT~\cite{yu2023unest} was proposed with a hierarchical transformer for segmenting kidney tumors and brain tumors.
For non-U-Net based models, DDTNet~\cite{zhang2022ddtnet} was designed for the detection and segmentation of breast cancer in histopathological images. Christ et~al.~\cite{christ2017automatic} employed cascaded fully convolutional neural networks (FCNN) for the automatic segmentation of liver and tumors. Other studies based on graph neural networks (GNNs)~\cite{patel2023multi} or transformer~\cite{zhou2023nnformer} also show great potential for accurate and effective tumor segmentation.

Although these techniques demonstrate remarkable segmentation performance, those fully supervised methods predominantly rely on a substantial amount of labeled data. However, pixel-level annotation is time- and cost-consuming, and it requires domain expert knowledge. Moreover, numerous variants of U-Net and non-U-Net architectures, particularly in the 3D medical imaging domain, often have a significantly large number of parameters. The increase in parameters not only raises computational complexity but also has the potential to hinder any practical implementations of these models in real-world situations.

\subsection{Semi-supervised medical image segmentation}
The primary challenge in medical image segmentation is the scarcity of labeled samples. This limitation has promoted the development of various SSL techniques. SSL techniques can be broadly classified into three categories. (1) Consistency regularization: Consistency regularization based methods aim to minimize the variance in predictions between a given unlabeled example and its perturbed counterpart.
For instance,
DTC~\cite{luo2021semi} employed a dual-task network to concurrently perform pixel-level classification and level set representation. URPC~\cite{luo2022semi} aimed to minimize the discrepancy between each pyramid prediction and their collective average, enhancing prediction reliability. 
CMMTA~\cite{cai2024class} developed class-aware mutual mixup with triple alignments for semi-supervised cross-domain segmentation.
CPC-SAM~\cite{miao2024cross} proposed a cross prompting consistency method with segment anything model for semi-supervised medical image segmentation.
(2) Self-training and pseudo-labeling: In self-training~\cite{bai2017semi}, pseudo labels are used to enhance the learning from unlabeled data. Chaitanya et~al.~\cite{chaitanya2023local} leveraged pseudo labels for unlabeled images by integrating a local contrastive loss to improve pixel-level feature learning in segmentation tasks.
STAMP~\cite{hasan2022stamp} tackled the issue of pseudo-labeling bias by adopting a self-training approach complemented by meta pseudo-labeling.
PLGCL~\cite{basak2023pseudo} introduced a semi-supervised patch-based contrastive learning framework for medical image segmentation that leverages pseudo labels to guide the learning process without relying on explicit pretext tasks.
MLRPL~\cite{su2024mutual} proposed a mutual learning framework that enhances pseudo-label reliability by comparing predictions between subnetworks, thereby improving segmentation performance.
BCP~\cite{bai2023bidirectional} enhanced the copy-paste concept in a bidirectional and symmetrical manner and reduced the distribution disparity between labeled and unlabeled data.
Recently, the effectiveness of BCP has attracted considerable research attention. For example,
SDCL~\cite{song2024sdcl} introduced a student discrepancy-informed correction learning framework, leveraging the disagreement between multiple student models to refine pseudo labels and improve segmentation performance.
MBCP~\cite{yu2024enhancing} proposed integrating Masked Autoencoders (MAE) into the BCP framework, introducing a masked image reconstruction task to improve the model's understanding of image structures.
DACNet~\cite{wang2025data} combined cross-pseudo labeling with strong and weak data augmentation strategies to enhance model performance and generalizability.
(3) Adversarial learning: Adversarial learning has drawn considerable attention in recent years. ACTS~\cite{lei2022semi} combined adversarial learning with consistency learning to explore a novel dynamic convolution based adversarial network for liver and skin lesion segmentation.

Nonetheless, these approaches are generally designed for segmenting larger organs. As a result, when employed for tumor segmentation, they often experience a decline in performance due to the distinctive attributes of tumors, such as their small size and uneven contrast. Moreover, they may neglect the significance of augmentation methods and the strategic incorporation of pseudo labels in the context of tumor segmentation. Finally, some recent methods~\cite{yu2024enhancing,wang2025data} incorporate augmentation strategies, they place excessive emphasis on the augmentation process, leading to high computational costs due to training with diffusion models and MAE, while lacking dynamic adaptation for tumors. These constraints motivate our approach. Our proposed solution dynamically employs a combination of weak and strong augmentation techniques and strategically utilizes pseudo labels for copy-paste supervision. This strategy aims to provide more informative training samples, thereby enhancing the robustness of the base network in tumor segmentation.

\begin{figure*}
  \centering
  \includegraphics[scale=0.47]{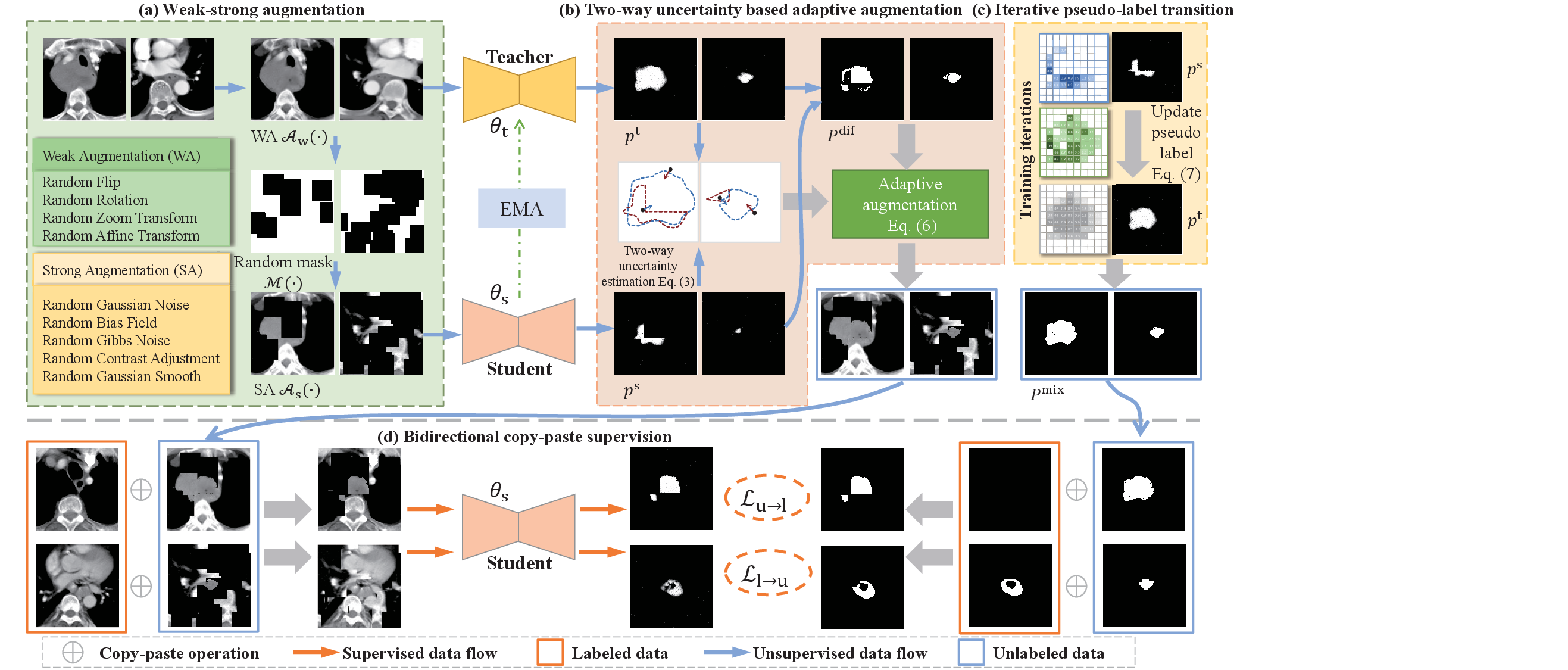}
  \caption{The overall architecture of iterative pseudo-labeling based adaptive copy-paste supervision (IPA-CP). (a) The weak-strong augmentation strategy. (b) Two-way uncertainty based adaptive augmentation. (c) Iterative pseudo-label transition. (d) Bidirectional copy-paste supervision. Typical methods of weak and strong augmentations are listed and random binary masks with different hole numbers and sizes are applied to the weakly augmented samples.
  }
  \label{fig:overview of framework}
\end{figure*}

\section{Methodology}
The overall workflow of the proposed IPA-CP method is given in Fig.~\ref{fig:overview of framework} with pseudo code given in Algorithm~\ref{code:IPACP}. The IPA-CP consists of two-way uncertainty based adaptive augmentation (TUAA), iterative pseudo-label transition (IPT), and bidirectional copy-paste (BCP) supervision modules. In the TUAA module, we first use a weak-strong augmentation strategy (Fig.~\ref{fig:overview of framework}(a)) to generate a pair of perturbed unlabeled samples and send them to the mean teacher architecture to obtain the predictions. Afterwards, predictions from teacher and student models are input into a two-way uncertainty estimation module (Fig.~\ref{fig:overview of framework}(b)) to quantify the uncertainty variance relative to each model. The uncertainty variances are included in an adaptive augmentation step, aiming to mitigate the potential negative impacts of strong augmentations on data distribution. The TUAA ensures the generation of diverse yet non-disruptive augmented unlabeled samples for training. In the IPT module, pseudo labels from the student model are adjusted to align more closely with those from the teacher model, as depicted in Fig.~\ref{fig:overview of framework}(c). Following the acquisition of unlabeled samples and the corresponding pseudo labels, the BCP module is integrated for bidirectional learning from both labeled and unlabeled datasets, as shown in Fig.~\ref{fig:overview of framework}(d).

\subsection{Two-way uncertainty based adaptive augmentation}
\subsubsection{Prediction disagreement generation}

Consistent with contemporary SSL methodologies~\cite{sohn2020fixmatch,chen2023boosting,wei2023towards}, we employ the mean teacher architecture to model prediction discrepancies. The weights $\theta_{\text{t}}$ of the teacher model are updated by the exponential moving average (EMA) weights $\theta_{\text{s}}$ of the student model $\mathcal{F}$.
Our objectives are twofold: (1) to encourage the student model to gain a more robust and generative learning capability, and (2) to amplify the prediction disagreement and introduce more variability, particularly in the context of small tumor data.

To better model prediction discrepancies, we employ both weak and strong augmentations on unlabeled samples. Mathematically, consider a labeled dataset $D_{\mathrm{l}} =\left\{(x_{i},y_{i})\right\}_{i=1}^{|D_{\mathrm{l}}|}$ and an unlabeled dataset $D_{\mathrm{u}} =\left\{(u_{i})\right\}_{i=1}^{|D_{\mathrm{u}}|}$, where $|D_{\mathrm{l}}|$ and $|D_{\mathrm{u}}|$ denote the numbers of labeled and unlabeled images, respectively, with $\left|D_{\mathrm{l}}\right| \ll\left|D_{\mathrm{u}}\right|$. Each image $x_{i}$ has a corresponding segmented ground truth $y_{i}$ in set $D_{\mathrm{l}}$. Set $D_{\mathrm{u}}$ comprises images used during training without annotations. For each unlabeled instance $u_i$, we generate a weakly augmented view using $\mathcal{A}_{\mathrm{w}}(\cdot)$ and a strongly augmented view with $\mathcal{A}_{\mathrm{s}}(\cdot)$.
As detailed in Table~\ref{table:augmentation}, the weak augmentation $\mathcal{A}_{\mathrm{w}}(\cdot)$ includes operations such as random flipping, rotation, zooming, and  affine transformation. In contrast, strong augmentations, aimed at creating large prediction disagreement through data perturbations, employ techniques such as adding Gaussian noise, bias field, Gibbs noise, contrast adjustment, and Gaussian smoothing. To further enhance disagreement and introduce variability in small tumor data, a random region mask $\mathcal{M}_{i}$ with different hole numbers and sizes is generated for $u_i$. Subsequently, a mixing step is applied between the strongly augmented sample and mask $\mathcal{M}_{i}$. The soft predictions are then mathematically computed as follows:
\begin{equation}
  \begin{split}
   & p_i^\text{s}=\mathcal{F}_{\theta_{\text{s}}}\left(\mathcal{M}_{i} \mathcal{A}_{\mathrm{s}}(\mathcal{A}_{\mathrm{w}}(u_i))\right),\\
   & p_i^\text{t}=\mathcal{F}_{\theta_{\text{t}}}\left(\mathcal{A}_{\mathrm{w}}(u_i)\right).
  \end{split}
    \label{eq:prediction}
\end{equation}
Here, $p_i^\text{s}$ represents the segmentation prediction generated by the student model while $p_i^\text{t}$ corresponds to that from the teacher model. Finally, the prediction disagreement $P^{\text{dif}}_{i}$ can be formulated as follows:
\begin{equation}
  \begin{split}
   & P^{\text{s}}_{i}=\argmax (p_i^\text{s}),  P^{\text{t}}_{i}=\argmax (p_i^\text{t}),\\
   & P^{\text{dif}}_{i}=||P^{\text{s}}_{i} - P^{\text{t}}_{i}||,
  \end{split}
    \label{eq:prediction_disagg}
\end{equation}
where $P^{\text{s}}_{i}$ and $P^{\text{t}}_{i}$ are pseudo labels.

\begin{table}
  \centering
  \begin{threeparttable}
    \caption{List of various weak-strong augmentations.}
    \label{table:augmentation}
    \renewcommand\tabcolsep{1.3pt}
    \begin{tabular}{l|l|l}
      \toprule
     Aug &  Methods &  Description \\ \hline
     \multirow{4}{*}{Weak}  & Flip& Randomly flip the image at spatial axis 0 with a probability of 0.3   \\
      &Rotation&Randomly rotate the image along the sagittal and coronal plane with a probability of 0.3 \\
      &Zoom transform&Randomly zoom in the image by [0.9, 1.1] with a probability of 0.3  \\
      &Affine transform&Randomly affine transform the image with a probability of 0.3	 \\\hline
      \multirow{6}{*}{Strong} &Gaussian noise &	Add Gaussian noise to the image with a probability of 0.5  \\
      &Bias field&Enhance the image by random bias field with a probability of 0.5	\\
      &Gibbs noise&	Add Gibbs noise to the image with a probability of 0.5 \\
      &Contrast adjustment&	Adjusts the contrast of the image by [1.2, 2] with a probability of 0.5\\
      &Gaussian smooth&	Smooth the image with a Gaussian kernel with a probability of 0.5\\
      \bottomrule
    \end{tabular}
  \end{threeparttable}
\end{table}

\subsubsection{Two-way uncertainty estimation}
In SSL tumor segmentation, the accurate assessment of uncertainty in unlabeled data is challenging due to two primary factors, the absence of precise ground truth (GT) labels, and the need to effectively capture subtle variances in predictions between teacher and student models. To address the two challenges, we design a two-way uncertainty estimation strategy to evaluate the inconsistency between the segmentation results generated by the two models.

Specifically, as illustrated in Fig.~\ref{fig:overview of framework}(b), for the $i$-th unlabeled sample, we initially derive the soft predictions $p_i^\text{s}$ and $p_i^\text{t}$ as outlined in Eq. (\ref{eq:prediction}). We then calculate the Kullback-Leibler (KL) divergence between these two predictions to quantify the variance:
\begin{equation}
  \begin{split}
    D_{\text{s}\rightarrow \text{t}}=\mathbb{E}\left[p_i^\text{s} \log (\frac{p_i^\text{s}}{p_i^\text{t}})\right], D_{\text{t}\rightarrow \text{s}}=\mathbb{E}\left[p_i^\text{t} \log(\frac{p_i^\text{t}}{p_i^\text{s}})\right],
  \end{split}
  \label{eq:uncertainty}
\end{equation}
here, $D_{\text{s}\rightarrow \text{t}}$ represents the divergence in probability distribution from the student model relative to the teacher model, while $D_{\text{t}\rightarrow \text{s}}$
indicates the divergence in the opposite direction. When the two models yield divergent class predictions, the approximated variance will show a high value. This elevated variance serves as an indicator of the model's uncertainty regarding the prediction.

\subsubsection{Adaptive augmentation}
To mitigate the potential adverse effects of strong augmentations on data distribution and segmentation performance, we develop an adaptive augmentation strategy for refining the augmentation of unlabeled samples. This approach utilizes quantitative two-way uncertainty estimations for each weakly and strongly augmented sample, aiming to highlight the uncertain regions through prediction disagreement. We initially implement a symmetric uncertainty thresholding approach for the segmentation results from both teacher and student models, focusing on high-uncertainty predictions. The computation process for the high-uncertainty score is detailed as below:
\begin{equation}
  \begin{split}
     \mu=& \mathbb{E}[\left(1-\mathbb{I}\left(\max\left(p_i^\text{t}\right) \geq \tau\right)\right) D_{\text{t}\rightarrow \text{s}}\\
     & +\left(1-\mathbb{I}\left(\max\left(p_i^\text{s}\right) \geq \tau\right)\right)D_{\text{s}\rightarrow \text{t}}],\\
  \end{split}
  \label{eq:adaptive_aug1}
\end{equation}
where $\mathbb{I}$ is the indicator function, and $\tau= 0.9$ represents a predefined threshold, used to filter out voxels with high uncertainty levels. In this way, samples exhibiting greater uncertainty among the teacher and student models are assigned a larger value of $\mu$. Conversely, samples with lesser uncertainty are identified by a smaller value of $\mu$.

After obtaining the high-uncertainty score, we further adaptively refine the weak-strong augmentation. The adaptive augmented version $\hat{u}_i$ of each unlabeled sample $u_i$ is then formulated as follows:
\begin{equation}
  \begin{split}
    & \hat{u}_i = (1-\mu)\mathcal{A}_{\mathrm{w}}(u_i)+\mu \mathcal{M}_i\mathcal{A}_{\mathrm{s}}(\mathcal{A}_{\mathrm{w}}(u_i)).\\
  \end{split}
  \label{eq:adaptive_aug2}
\end{equation}
Consequently, $\hat{u}_i$ closely resembles the weakly augmented version of $u_i$ in cases of low uncertainty, while in other scenarios, $\hat{u}_i$ aligns more closely with the masked strongly augmented version of $u_i$. This approach renders the augmentation process both dynamically and informatively.

To model discrepancies in predictions, we also incorporate variations caused by the random binary mask $\mathcal{M}_i$. Consequently, the unlabeled sample $\hat{u}^{\text{a}}_i$ is augmented as follows:
\begin{equation}
  \begin{split}
    & \hat{u}^{\text{a}}_i = (1-P^{\text{dif}})\hat{u}_i+P^{\text{dif}} \mathcal{M}_i\mathcal{A}_{\mathrm{s}}(\mathcal{A}_{\mathrm{w}}\left(u_i\right)).
  \end{split}
  \label{eq:adaptive_aug3}
\end{equation}

In this way, the diversity of the tumor representation is continuously enhanced.

\subsection{Iterative pseudo-label transition}
Pseudo-labeling has been established as a straightforward yet effective technique for SSL in the medical field, as noted in~\cite{zheng2021rectifying,lyu2022pseudo,chaitanya2023local}. Most advanced pseudo-labeling methods emphasize developing sampling strategies that prioritize high-confidence pseudo labels, aligning with task-specific requirements. However, these methods may overlook the refinement of pseudo labels in the teacher-student architecture for unlabeled data, potentially leading to overconfident predictions in areas where the student model's predictions are ambiguous. This is particularly the case in tumor segmentation, where distinguishing the foreground from the background can be exceptionally challenging, making it difficult to select reliable pseudo labels. Moreover, the teacher model does not always guarantee better results than the student model on unlabeled data~\cite{jin2024inter}. To address this, we propose the IPT strategy to fine-tune the soft pseudo labels (i.e., $p_i^\text{t}$ and $p_i^\text{s}$) derived from teacher and student models. This fine-tuning process involves moving the perturbed predictions of the student model closer to the more stable predictions of the teacher model. The pseudo-label target is updated as follows:
\begin{equation}
  \begin{split}
    & P^{\text{mix}}_{i}=\argmax (\frac{e}{e+1} p_i^\text{t}+\frac{1}{e+1} p_i^\text{s}),
  \end{split}
  \label{eq:pseudo_label_transform}
\end{equation}
where $e$ is the $e$-th iteration of the total $E$ training iterations. In practical implementation, the epoch number is denoted as $e$, and $E$ is the total training epochs.
As a result, more robust yet diverse pseudo labels are generated for unlabeled samples.

\begin{algorithm}[h]
  \caption{Iterative pseudo-labeling based adaptive copy-paste supervision}
  \label{code:IPACP}
  \begin{algorithmic}[1]
    \Require
    $\mathcal{B}_{\text{x}}=\left\{\left(x_i, y_i\right)\right\}_{i=1}^{\left|\mathcal{B}_{\text{x}}\right|}$: labeled batch;
    $\mathcal{B}_{\text{u}}=\left\{u_i\right\}_{i=1}^{\left|\mathcal{B}_{\text{u}}\right|}\left(\left|\mathcal{B}_{\text{x}}\right|=\left|\mathcal{B}_{\text{u}}\right|\right)$: unlabeled batch; $\tau$: hyperparameter threshold;
    $\mathcal{A}_{\mathrm{w}}(\cdot)$: weak augmentation;
    $\mathcal{A}_{\mathrm{s}}(\cdot)$: strong augmentation;
    $\mathcal{M}$: randomly generated mask batch; $E$: total number of training iterations
    \Ensure
    Optimal $\theta_{\text{s}}$ 
    \For{iteration $e$ = 1 to $E$}
    \State Generate a minibatch of unlabeled data $\mathcal{B}_{\text{u}}$
    \For{each $u_i \in \mathcal{B}_{\text{u}}$}
    \State Augment $u_i$ using $\mathcal{A}_{\mathrm{w}}(\cdot)$ and $\mathcal{A}_{\mathrm{s}}(\cdot)$
    \State Compute soft prediction pair ($p_i^\text{t}$, $p_i^\text{s}$) using Eq.~(\ref{eq:prediction}) 
    \State Generate prediction disagreement $P^{\text{dif}}_{i}$ via Eq.~(\ref{eq:prediction_disagg})
    \State Estimate the two-way uncertainties ($D_{\text{s}\rightarrow \text{t}}$, $D_{\text{t}\rightarrow \text{s}})$ using Eq.~(\ref{eq:uncertainty})
    \State Calculate high-uncertainty score $\mu$ via Eq.~(\ref{eq:adaptive_aug1})
    \State Adaptively augment $u_i$ using Eq.~(\ref{eq:adaptive_aug2}) and Eq.~(\ref{eq:adaptive_aug3})
    \State Iterative pseudo-label transition by Eq.~(\ref{eq:pseudo_label_transform})
    \EndFor
    \State Generate a minibatch of labeled data $\mathcal{B}_{\text{x}}$
    \For{each $i \in [0,\left|\mathcal{B}_{\text{x}}\right|/2)$}
    \State Bidirectional copy-paste the labeled and unlabeled sample using Eq.~(\ref{eq:bcp_super1})
    \State Bidirectional copy-paste the ground truth and pseudo label using Eq.~(\ref{eq:bcp_super2})
    \State Calculate supervised loss using Eq.~(\ref{eq:loss_func})
    \EndFor
    \State Update weight $\theta_{\text{s}}$ of the student model
    \State Update weight $\theta_{\text{t}}$ using the EMA strategy with decay rate $\alpha$
    \EndFor
  \end{algorithmic}
\end{algorithm}

\subsection{Bidirectional copy-paste supervision}
To synergistically leverage knowledge from both labeled and unlabeled data, we integrate the BCP method as described in~\cite{bai2023bidirectional}, facilitating the final image composition and pseudo label mixup. Consider a labeled minibatch $\mathcal{B}_{\text{x}}=\left\{(x_{i},y_{i})\right\}_{i=1}^{|\mathcal{B}_{\text{x}}|}$ and an unlabeled minibatch $\mathcal{B}_{\text{u}}=\left\{u_i\right\}_{i=1}^{\left|\mathcal{B}_{\text{u}}\right|}$, with the randomly generated binary masks $\mathcal{M}$ and equal batch sizes ($|\mathcal{B}_{\text{x}}|=|\mathcal{B}_{\text{u}}|$). The binary mask differentiates voxels from the foreground (0) or background (1). We implement a bidirectional method: first, by copying the background from the $i$-th labeled sample and pasting it onto the $i$-th unlabeled sample in a minibatch, we create a mixed sample denoted as $X_{\text{l}\rightarrow \text{u}}$. Conversely, we copy the background of the $j$-th unlabeled sample and paste it onto the $j$-th labeled sample, resulting in a mixed sample represented as $X_{\text{u}\rightarrow \text{l}}$. These processes can be represented in the following equations:
\begin{equation}
  \begin{split}
    & X_{\text{l}\rightarrow \text{u}}=\mathcal{M}_{i}x_i+(1-\mathcal{M}_{i})\hat{u}^{\text{a}}_i,\\
    & X_{\text{u}\rightarrow \text{l}}=\mathcal{M}_{j}\hat{u}^{\text{a}}_{j}+(1-\mathcal{M}_{j})x_{j}.
  \end{split}
  \label{eq:bcp_super1}
\end{equation}

Correspondingly, for training the student network, we generate supervisory signals using the BCP method. Leveraging the IPT technique, we treat $P^{\text{mix}}_{i}$ as the refined ground truth for the unlabeled data. The ground truth for the labeled images is processed in a similar manner, as detailed in Eq.~(\ref{eq:bcp_super1}).
\begin{equation}
  \begin{split}
    & Y_{\text{l}\rightarrow \text{u}}=\mathcal{M}_{i}y_i+(1-\mathcal{M}_{i})P^{\text{mix}}_{i},\\
    & Y_{\text{u}\rightarrow \text{l}}=\mathcal{M}_{j}P^{\text{mix}}_{j}+(1-\mathcal{M}_{j})y_j,
  \end{split}
  \label{eq:bcp_super2}
\end{equation}
with $j={i+\frac{\left|\mathcal{B}_{\text{u}}\right|}{2}}$. It is important to note that in our implementation, we have omitted the ``largest connected component'' strategy, which is a standard component of BCP. Unlike large anatomical structures, where the largest component is typically the correct one, small organs, and tumors may appear as multiple fragments due to imaging artifacts or partial visibility. These small structures can consist of multiple disconnected yet clinically relevant components. Applying such a strategy may mistakenly remove these smaller structures, leading to the loss of essential information and making it less effective for segmenting numerous small tumors.
To this end, we obtain ($X_{\text{l}\rightarrow \text{u}}$, $Y_{\text{l}\rightarrow \text{u}}$) and ($X_{\text{u}\rightarrow \text{l}}$, $Y_{\text{u}\rightarrow \text{l}}$) pairs for supervised training.

\subsection{Loss function}
As the student model is trained in a supervised manner, the choice of a loss function can be flexible. To be consistent with common practices in medical image segmentation, we employ both cross-entropy loss (CE) and Dice loss. Therefore, the formulation of the calculation is as follows:
\begin{equation}
  \begin{split}
    \mathcal{L}_{\text{l}\rightarrow \text{u}}= & \operatorname{Dice}\left(\mathcal{F}_{\theta_{\text{s}}}(X_{\text{l}\rightarrow \text{u}}),Y_{\text{l}\rightarrow \text{u}}\right)
    +\operatorname{CE}\left(\mathcal{F}_{\theta_{\text{s}}}(X_{\text{l}\rightarrow \text{u}}),Y_{\text{l}\rightarrow \text{u}}\right),\\
    \mathcal{L}_{\text{u}\rightarrow \text{l}}=& \operatorname{Dice}\left(\mathcal{F}_{\theta_{\text{s}}}(X_{\text{u}\rightarrow \text{l}}),Y_{\text{u}\rightarrow \text{l}}\right)
    +\operatorname{CE}\left(\mathcal{F}_{\theta_{\text{s}}}(X_{\text{u}\rightarrow \text{l}}),Y_{\text{u}\rightarrow \text{l}}\right).\\
  \end{split}
  \label{eq:loss_func}
\end{equation}
\section{Data collection and preprocessing}
\subsubsection{FSD} We acquired thoracic CT scans of 558 fistula esophageal cancer patients from Shandong Cancer Hospital. This study was approved by the Shandong Cancer Hospital and Institute Ethics Committee (2022001007). These scans feature a Hounsfield unit (HU) range of -1,024 to 3,071 and voxel sizes varying between $0.5429 \times 0.5429 \times 3$ mm$^3$ and $1.2519 \times 1.2519 \times 5$ mm$^3$. The resolution of the thoracic CT scans spans from $512 \times 512 \times 71$ to $512 \times 512 \times 114$ voxels. For accurate analysis, esophageal tumors were manually delineated and verified by two radiotherapists. Tumor volumes vary from $16 \times 16 \times 3$ to $99 \times 141 \times 40$ voxels. We divided the dataset into 390 training and 168 testing samples. From the training set, 30 samples were randomly chosen for validation purposes.

\subsubsection{MSD} We further evaluated the generalization ability of our proposed architecture using the Liver Tumor Segmentation Challenge (LiTS) dataset~\cite{antonelli2022medical}, comprising 200 CT scans. This dataset is one of the most popular datasets for liver tumor segmentation and it is partitioned into 130 scans for training and 70 scans for testing, all with a uniform in-plane resolution of 512 $\times$ 512 but with varying numbers of axial slices. However, the ground truth for the test data remains unreleased. The tumor count in these scans ranges from 0 to 12, with sizes varying from 38 mm$^3$ to 1,231 mm$^3$. After removing scans without tumors, we retained 116 out of the 130 scans for our analysis. These were then randomly split into three subsets: 100 for training, 8 for validation, and 8 for testing.

In the image preprocessing phase, we truncated the intensity values of all scans to a range of [-200, 300] HU for the FSD dataset and [-100, 200] HU for the MSD dataset to eliminate irrelevant details from the scans. For the FSD dataset, all images were resampled to a uniform resolution of $0.8\times0.8\times5$ mm$^3$. Similarly, for the MSD dataset, the images were resampled to a standard resolution of $2\times2\times2$ mm$^3$. Subsequently, we applied min-max normalization to obtain the final preprocessed volumes.

\section{Experiments and results}
\subsection{Implementation details and evaluation measures}
IPA-CP is implemented in PyTorch and the MONAI framework using an NVIDIA RTX 3090 graphic card. We set the batch size as 2 and the training patch size as $160\times160\times80$ for esophageal tumors and $112\times 112 \times 64$ for liver tumors, respectively. The Adam optimizer is used with a polynomial learning rate policy, where the initial learning rate $\num{2.5e-4}$ is multiplied by $\left(1-\frac{iter}{total\_{iter}}\right)^{power}$ with $power$ being 0.9. The total number of training iterations is set to 6,000 for esophageal tumors and 42,000 for liver tumors. The EMA decay weight factor $\alpha$ is set to 0.99. We use 3D U-Net as our base network. All the performances are obtained from the average of 3 runs.

\begin{sidewaystable*}
  \centering
  \begin{threeparttable}
    \caption{Experimental results on FSD and MSD using our IPA-CP and state-of-the-art SSL methods with different percentages of labeled data. Note that 10\%(36/10) denotes 10\% labeled data and the corresponding number of labeled data is 36/10 for the FSD/MSD dataset.}
    \label{table:method_comparison}
    \renewcommand\tabcolsep{2.5pt}
    \begin{tabular}{c|cc|ccccc|ccccc}
      \toprule
      \multirow{2}{*}{Method} &\multicolumn{2}{c|}{Data used}   &  \multicolumn{5}{c|}{FSD}   & \multicolumn{5}{c}{MSD}  \\ \cline{2-13}
      & Labeled&Unlabeled&  Dice~(\%)$\uparrow$   &  JA~(\%)$\uparrow$   & RMSE~(\%)$\downarrow$  & HD~(mm)$\downarrow$  &ASD~(mm)$\downarrow$ &   Dice~(\%)$\uparrow$&JA~(\%)$\uparrow$   &  RMSE~(\%)$\downarrow$  &HD~(mm)$\downarrow$ &ASD~(mm)$\downarrow$\\  \hline
      U-Net& 10\%&0&  63.77(.29) &  49.39(.45) & 6.66(.20)  &13.63(2.71) &4.78(1.18) &33.73(1.54)&24.85(1.54)&6.32(.02)&33.89(7.11)&20.83(2.95) \\
      U-Net& 20\%&0& 67.50(.44) &  53.62(.46) & 6.21(.09)&9.68(.83)   &3.51(.49)   &36.21(1.31)&26.46(1.41)&	5.87(.04)&25.43(7.79)&15.09(6.74)\\
      U-Net& 100\%& 0 &74.72(.66)& 61.86(.65) &5.46(.10)&5.95(.53)&	1.86(.23)&54.19(3.02) &43.55(2.41) &5.49(.10) &15.66(2.65)&	6.77(.71) \\ \hline
      MT (NeurIPS'17)& \multirow{11}{*}{\makecell[c]{10\% \\(36/10)} } & \multirow{11}{*}{\makecell[c]{90\%\\(324/90) }}& 64.83(.23)& 50.30(.17) &	6.60(.11)	&11.60(.96)& 4.14(.46)	&35.81(.47) &	26.95(.44) &	6.28(.07) &	34.38(10.06)& 	22.03(3.40) \\
      ICT (AAAI'19)   && &  64.94(.19)&50.34(.14)&6.55(.03) &	11.42(.85)&	3.98(.34)	 &35.41(.54)&26.30(.37)&6.17(.02)& 	34.88(4.02)&	19.39(1.63)  \\
      CCT (CVPR'20) &&  &66.84(.44)&52.68(.40)	& 6.27(.09)&	10.82(.97)	&3.76(.28)  & 35.27(1.25)	&26.74(1.02)&	6.07(.09)& 32.16(7.21)&	17.81(5.94) \\
      CPS (CVPR'21) && & 66.93(.49) 	&52.71(.47)&	6.21(.03) &	9.65(1.11) &	3.31(.36)&34.96(3.56) &	26.25(3.14) &6.20(.09)	&	36.62(2.31)& 	22.33(3.14)  \\
      RD (NeurIPS'21) &&  &64.97(.98)&	50.54(.87)&	6.42(.10)& 9.98(.81)&	3.45(.57)	 &35.15(1.85)&26.27(1.45)&6.22(.06)&	36.18(12.10)&	21.02(6.53)  \\
      URPC (MIA'22) && & 65.24(1.10)& 50.92(1.05)&	6.47(.12)& 10.45(.47)&	3.26(.13)	&34.43(.83)&	25.90(.75)&6.12(.21)&27.53(12.25)&16.05(6.43)\\
      ACL (TMI'23) && &67.19(.24) &	52.96(.32) 	&6.21(.04)&  8.10(.23)& 	2.46(.12)  &22.21(2.13)&15.72(1.79) &6.93(.25) &29.66(13.56)&16.21(10.56) \\
      BCP (CVPR'23) && &67.76(.17)	&52.89(.20)	&6.96(.04)& 12.14(.81)&	3.95(.20)   &40.11(1.61)&30.78(1.01)&	\textbf{\color{red}{5.51(.12)}} &	28.61(7.04)&16.10(5.89)\\
      MCF (CVPR'23)&& &66.68(.78) &52.38(.68)& 6.31(.02) &10.55(.29)& 	3.43(.07) &28.10(.74) &	19.45(1.07) &6.26(.23) &\textbf{\color{red}{25.19(3.08)}}&	\textbf{\color{blue}{13.13(4.78)}} 	 \\
      MBCP (IJMLC'24)& & &67.50(.51) &53.01(.92)&6.52(.29)&11.46(1.21)& 3.82(.35)& 43.28(.87)&32.55(2.03) &5.63(.42) &33.18(6.98)&14.57(3.18) \\
      MLRPL (MIA'24)&& &67.08(.59) &52.30(.71)&6.51(.11)&11.10(.37)& 3.61(.21)& \textbf{\color{red}{46.21(.64)}}&\textbf{\color{red}{36.88(.66)}} &6.66(.58) &33.58(2.98)&21.43(2.36) 	 \\
      SDCL (MICCAI'24)&& & \textbf{\color{red}{70.53(.11)}}&\textbf{\color{red}{57.18(.11)}}&\textbf{\color{red}{5.78(.04)}}&\textbf{\color{blue}{7.11(.10)}}&\textbf{\color{red}{2.15(.17)}}& 37.88(1.67)&28.88(1.42) &5.55(.09) &\textbf{\color{blue}{22.25(8.72)}}&\textbf{\color{red}{13.23(7.58)}} 	 \\
      DACNet (PR'25) && &68.30(1.26) &54.63(1.15)&6.02(.11) &8.07(.60)&2.56(.30)& 34.00(2.19) &25.66(2.25) &5.97(2.25) &25.87(11.09)&13.87(6.16) 	 \\
      IPA-CP && &\textbf{\color{blue}{71.72(.57)}} &	\textbf{\color{blue}{57.66(.54)}} &	\textbf{\color{blue}{5.77(.03)}}	&	\textbf{\color{red}{7.29(.36)}} &	\textbf{\color{blue}{2.05(.12)}}  &\textbf{\color{blue}{46.61(4.51)}}& \textbf{\color{blue}{37.34(3.23)}}& 	\textbf{\color{blue}{5.15(.42)}}  &30.21(5.89) &	17.61(3.77)\\  \hline
      MT (NeurIPS'17)& \multirow{11}{*}{\makecell[c]{20\% \\(72/20)}} & \multirow{11}{*}{\makecell[c]{80\% \\(288/80)}}&67.63(.47) &53.81(.44) &	6.11(.12)	&	8.78(.81) &	2.89(.59) &38.18(4.00) &28.26(2.83)& 5.79(.09) &22.27(3.24)&13.40(4.64)  \\
      ICT (AAAI'19) && &  67.88(.43)&54.01(.50)&	6.16(.14)&  9.48(.90)&	3.19(.36)&39.26(1.51)& 29.66(2.06)&	5.70(.15)&		29.79(6.72)&	16.47(8.19)  \\
      CCT (CVPR'20)  &&  &67.91(.40)&54.31(.21)&6.06(.10)& 8.94(1.37)&	3.06(.73) & 40.12(3.64)& 31.04(3.58)&\textbf{\color{blue}{4.65(1.64)}}&33.70(8.70)&	19.49(4.87) \\
      CPS (CVPR'21)  && & 68.00(.79) &	54.17(1.12) &	6.25(.18) &	9.37(1.77) &	3.29(.86) & 38.04(2.42)& 27.44(1.13) &	5.82(.05) &	23.90(7.01)&13.38(4.47) \\
      RD (NeurIPS'21)&&  &68.35(.94)&54.38(.82)&	6.17(.02)& 8.54(.53)&	2.84(.42)     &40.85(.50)&31.21(.10)&5.58(.12)&	30.70(1.28)&	17.73(2.69)\\
      URPC (MIA'22) && & 67.83(.49)&	54.27(.47)&	5.94(.13)& 8.06(.23)&	2.35(.19)  &37.13(2.84)& 28.90(2.58)&	5.65(.20)	&\textbf{\color{blue}{15.84(9.56)}}&	10.08(9.06)  \\
      ACL (TMI'23) && &69.92(.13) &56.13(.14) &	6.00(.02)& 7.10(.14)& 2.22(.10)  &24.12(2.53)& 17.15(2.44) &6.29(.20)&29.83(6.53)&	14.42(5.89)  \\
      BCP (CVPR'23) && &71.96(.28)&	57.75(.35)&	6.29(.05)& 9.21(.17)	&3.00(.11)& 43.02(1.35) &32.72(1.79)&5.57(.36)&26.25(6.27)&	14.86(4.86)   \\
      MCF (CVPR'23)&& &70.64(.65)& 57.01(.58) &	5.88(.07)&	7.99(.37)& 	2.65(.30) &32.33(8.54) &23.09(7.41) &6.20(.39) &30.07(18.36)&11.91(9.62)  \\
      MBCP (IJMLC'24) && &71.18(.70)&56.80(.84)&6.37(.14)&9.15(0.72)&2.95(.23)& 46.18(1.69)&34.45(1.43) &5.50(.15) &26.60(6.04)&12.39(3.57) \\
      MLRPL (MIA'24) && &69.24(.09) &55.34(.23)&6.21(.09)  &9.05(.31)&3.05(09)&47.05(1.63) &\textbf{\color{red}{37.55(1.17)}} &4.97(.04) &\textbf{\color{red}{16.90(6.83)}}&\textbf{\color{blue}{7.60(4.31)}} \\
      SDCL (MICCAI'24) && &\textbf{\color{red}{73.19(.97)}} &\textbf{\color{red}{59.93(1.15)}}& \textbf{\color{red}{5.63(.10)}} &\textbf{\color{blue}{6.58(.40)}}& \textbf{\color{red}{1.89(.25)}}& \textbf{\color{red}{43.72(2.68)}}&33.28(2.99) &5.42(.23) &16.95(4.48)&\textbf{\color{red}{7.79(1.88)}} 	 \\
      DACNet (PR'25) && &70.85(.93) &57.31(.62)&5.82(.13) &7.28(.17)&2.21(.16)& 41.23(3.48)&30.81(2.80) &5.73(.08) &23.40(4.57)&13.33(2.05)\\
      IPA-CP && &\textbf{\color{blue}{73.78(.48)}}&\textbf{\color{blue}{59.93(.66)}}&\textbf{\color{blue}{5.62(.10)}}&\textbf{\color{red}{6.74(.30)}}&\textbf{\color{blue}{1.87(.13)}} &\textbf{\color{blue}{51.84(1.05)}} &	\textbf{\color{blue}{42.14(.43)}} &\textbf{\color{red}{4.85(.28)}} & 24.55(7.50)&12.76(2.63)	\\
      \bottomrule
    \end{tabular}
    \begin{tablenotes}[flushleft]
      \footnotesize
      \item \textit{Note:} The best responses are highlighted in {\textbf{\color{blue}{blue}}}, and the second best responses are in {\color{red}{\textbf{red}}}. Results are given in the Mean(Std) format.
      \end{tablenotes}
  \end{threeparttable}
\end{sidewaystable*}

We assess and compare the performance of various segmentation methods on the test dataset, employing quantitative metrics such as dice coefficient (Dice), Jaccard index (JA), 95\% Hausdorff distance (HD), root mean squared error (RMSE), and average surface distance (ASD).
\subsection{Quantitative evaluation and comparison}
We compare our method with several state-of-the-art semi-supervised segmentation approaches in the medical imaging domain. We select consistency regularization-based methods, including mean teacher (MT)~\cite{tarvainen2017mean}, interpolation consistency training (ICT)~\cite{verma2019interpolation}, cross-consistency training (CCT)~\cite{ouali2020semi}, uncertainty rectified pyramid consistency (URPC)~\cite{luo2022semi}, and regularized dropout (RD)~\cite{wu2021r}. Additionally, we consider self-training and pseudo-labeling based approaches, such as cross pseudo supervision (CPS)~\cite{chen2021semi}, mutual correction framework (MCF)~\cite{wang2023mcf}, mutual learning with reliable pseudo label (MLRPL)~\cite{su2024mutual}, bidirectional copy-paste (BCP)~\cite{bai2023bidirectional}, masked autoencoders based bidirectional copy-paste framework (MBCP)~\cite{yu2024enhancing}, students discrepancy-informed correction learning (SDCL)~\cite{song2024sdcl} and dual attention-guided consistency network (DACNet)~\cite{wang2025data}. Furthermore, we compare our method with adversarial methods, such as adversarial consistency learning (ACL)~\cite{lei2022semi}. Notably, BCP-based methods include BCP, MBCP, SDCL, and MACNet. All the models are trained from scratch to ensure the development of an unbiased learning process specific to our datasets. We employ two typical semi-supervised experimental settings~\cite{luo2021semi,wu2022mutual}, i.e., training with 10\%/20\% labeled data and 90\%/80\% unlabeled data, across all methods for fair comparison.
\subsubsection{FSD} As given in Table~\ref{table:method_comparison}, our IPA-CP demonstrates superior performance across all five evaluation metrics at both 10\% and 20\% labeled data ratios, showing the efficacy of our proposed framework. Notably, in scenarios with only 10\% labeled data, IPA-CP outperforms the second-best model, exhibiting a 7.95\% increase in Dice score compared to the baseline. Furthermore, it is noteworthy that the overall performance of our model is close to that achieved by fully supervised learning using the 100\% labeled dataset. These results underscore the effectiveness of IPA-CP in segmenting esophageal tumors.

\subsubsection{MSD} We further validate IPA-CP on the publicly available dataset, MSD, to demonstrate the generalization capability and superiority of our method. IPA-CP shows competitive results compared with other state-of-the-art approaches in Dice, JA, and RMSE metrics, especially in terms of Dice and JA. Notably, methods designed for large organ segmentation, such as SDCL and DACNet, perform well on the FSD dataset but fail when applied to tumor segmentation involving numerous small tumors. The findings presented in Table~\ref{table:method_comparison} provide evidence of IPA-CP's generalization capability and effectiveness on datasets with small tumors.

Several observations can be drawn from the results presented in Table~\ref{table:method_comparison}. 
(1) Our proposed IPA-CP method consistently achieves the highest Dice and JA scores across both FSD and MSD datasets, demonstrating its superiority in leveraging unlabeled data for semi-supervised tumor segmentation.
(2) The impact of labeled data proportions on segmentation performance is evident. As expected, increasing the proportion of labeled data leads to overall improvements across all methods.
(3) IPA-CP outperforms BCP-based approaches across nearly all metrics. While SDCL performs comparably to IPA-CP on FSD, it falls significantly behind when segmenting small tumors on MSD. We attribute this gap to the effectiveness of our two-way uncertainty-based adaptive augmentation strategy, which focuses on small boundaries.
(4) IPA-CP demonstrates robust performance in both large and small tumor datasets. While most SSL methods (e.g., SDCL and DACNet, which are designed for large organ segmentation) struggle with the MSD dataset, which contains smaller tumors, IPA-CP still achieves a significant improvement among all methods. This suggests that our approach is particularly well-suited for addressing the challenges of small tumor segmentation, a critical limitation of many existing SSL frameworks.

In summary, the results demonstrate that IPA-CP significantly outperforms existing SSL methods in tumor segmentation, particularly in low-label scenarios. Its ability to effectively integrate labeled and unlabeled data, refine pseudo labels, and enhance segmentation robustness adaptively makes it a promising approach for real-world medical imaging applications.

\begin{figure*}
  \centering
  \includegraphics[scale=0.38]{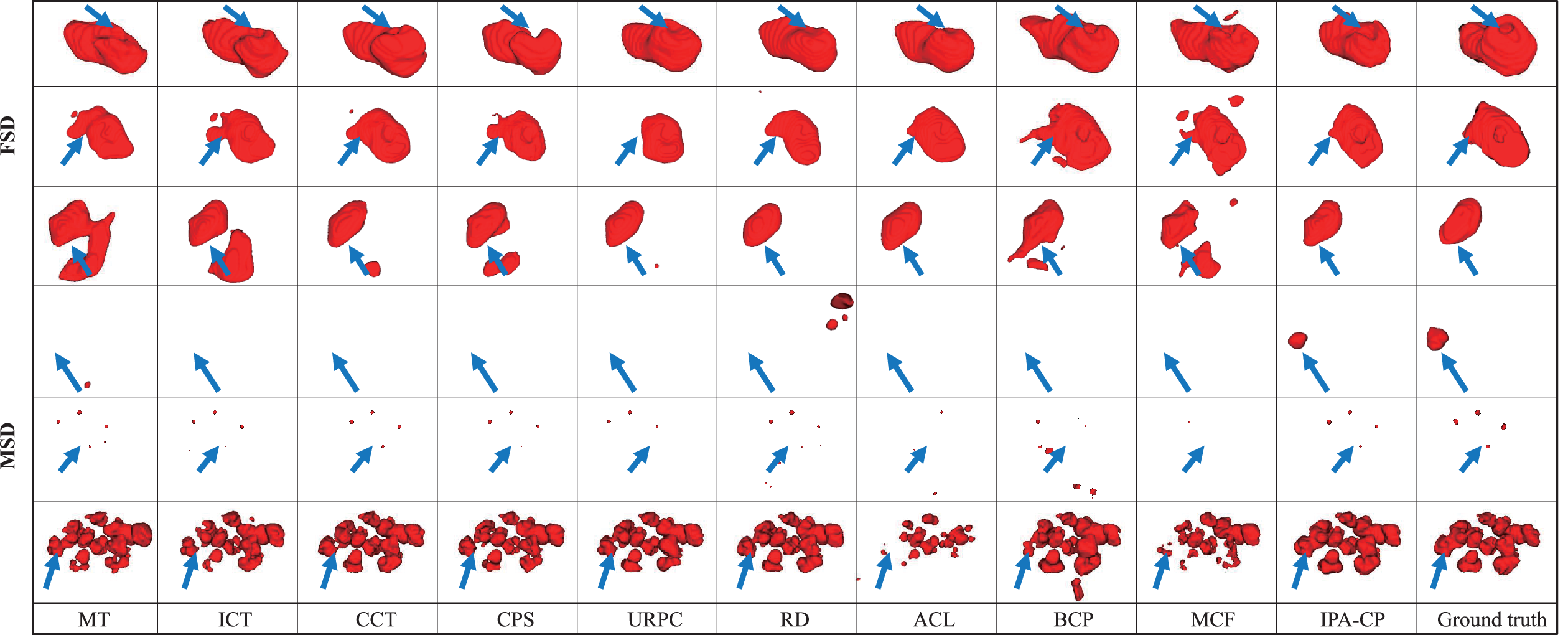}
  \caption{Qualitative results of three cases from the FSD dataset and three cases from the MSD dataset. The columns, arranged from left to right, present the 3D tumor segmentation results generated by typical state-of-the-art SSL methods on 10\% labeled data and 90\% unlabeled data, with the last column showing the ground truth.
  }
  \label{fig:ssl_res}
\end{figure*}

\subsection{Qualitative comparison}
To intuitively demonstrate the tumor segmentation ability endowed by IPA-CP, we conduct qualitative experiments on the FSD and MSD datasets, with visualizations of typical segmentation results presented in Fig.~\ref{fig:ssl_res}.
Specifically, for relatively large tumors, as exemplified by an esophageal tumor case in the second row, IPA-CP aligns predictions closely with the ground truth, whereas other methods either produce numerous false positives or miss key tumor regions.
Regarding small tumors, in the fourth row of the figure, IPA-CP successfully identifies small-sized tumors on the MSD dataset, a task at which other SSL methods fail. The fifth row further highlights the ability of IPA-CP to segment small tumors, as other methods also exhibit inferior performance with incorrect voxel predictions.
Overall, our IPA-CP method exhibits more accurate and detailed tumor segmentation, while other methods tend to incorrectly segment non-target regions or fail to identify all tumors.
We attribute this primarily to the presence of numerous ambiguous tumor voxels in CT images, which vary significantly in size and tend to confound models that are originally designed for larger organs.

\begin{table*}
  \centering
  \begin{threeparttable}
    \caption{Segmentation performance in the ablation studies on the FSD/MSD dataset using 10\%(36/10) labeled data.}
    \label{table:ablation_study}
    \renewcommand\tabcolsep{1.5pt}
    \begin{tabular}{ccccc|cccc|cccc}
      \toprule
      \multirow{2}{*}{WSA} & \multirow{2}{*}{BCP}& \multirow{2}{*}{TUE}& \multirow{2}{*}{IPT} & \multirow{2}{*}{PDI}   &  \multicolumn{4}{c|}{FSD}   & \multicolumn{4}{c}{MSD} \\\cline{6-13}
       & & & & & Dice~(\%)$\uparrow$  & JA~(\%)$\uparrow$   & RMSE~(\%)$\downarrow$ & HD~(mm)$\downarrow$   &  Dice~(\%)$\uparrow$  & JA~(\%)$\uparrow$   & RMSE~(\%)$\downarrow$ & HD~(mm)$\downarrow$   \\ \hline
       &&&& &63.77 &  49.39 & 6.66   &13.63  & 33.73&24.85&6.32&33.89  \\
      \boldsymbol{$\surd$}&&& & & 65.51&51.11&6.46&11.09&35.10&26.33&6.30&	33.03 \\
      \boldsymbol{$\surd$}  & \boldsymbol{$\surd$} &  & &  &70.61&56.58 &5.82 &7.63  &37.23 &28.09 &5.72& 23.36 \\
      \boldsymbol{$\surd$} & & \boldsymbol{$\surd$} &  &  & 68.88&54.48 &6.07 &6.98  &36.84&	27.71&5.73&27.93	 \\
      \boldsymbol{$\surd$} &\boldsymbol{$\surd$} &\boldsymbol{$\surd$} &  &  &71.34 &57.24&5.84&7.11 &43.98 &34.15&5.33&21.37  \\
      \boldsymbol{$\surd$}  &\boldsymbol{$\surd$} &  &    \boldsymbol{$\surd$}  &&71.10&56.86&5.88 &7.44  &38.22&	28.72&5.82&30.01\\
      \boldsymbol{$\surd$} &\boldsymbol{$\surd$} & \boldsymbol{$\surd$} &    \boldsymbol{$\surd$}  & \boldsymbol{$\surd$}& 71.72 &	57.66 &	5.77	&	7.29 &	46.61& 37.34& 	5.15  &	30.21 \\
      \bottomrule
    \end{tabular}
  \end{threeparttable}
\end{table*}

\subsection{Ablation study}
We perform ablation study to demonstrate the contribution of each component within the proposed framework, which include the weak-strong augmentation strategy under the mean teacher architecture (WSA), two-way uncertainty estimation (TUE, as detailed in Eq.~(\ref{eq:adaptive_aug2})), prediction disagreement incorporation (PDI, as outlined in Eq.~(\ref{eq:adaptive_aug3})), iterative pseudo-label transition (IPT, as described in Eq.~(\ref{eq:pseudo_label_transform})), and bidirectional copy-paste (BCP, as described in Eqs.~(\ref{eq:bcp_super1}) and~(\ref{eq:bcp_super2})). As indicated in Table~\ref{table:ablation_study}, the ablation studies are conducted using 10\%/90\% labeled/unlabeled samples on the FSD/MSD dataset.
\subsubsection{FSD} The ablation study on the FSD dataset highlights the contributions of each component in IPA-CP to segmentation performance. The baseline model equipped with the WSA strategy achieves a Dice score of 65.51\%, demonstrating the fundamental impact of weak and strong augmentation strategies. Adding BCP further enhances segmentation, increasing the Dice score to 70.61\% and reducing HD from 11.09 mm to 7.63 mm, confirming that BCP significantly improves training sample diversity for large tumors.
Introducing TUE while removing BCP also boosts performance, achieving a Dice score of 68.88\% by selectively enhancing augmentation based on uncertainty guidance. This result confirms the effectiveness of our TUE strategy. Adding IPT and PDI further improves performance, achieving the best overall results with a Dice score of 71.72\% and HD reduced to 7.29 mm, highlighting the importance of leveraging model disagreement.
The steady improvements across different settings confirm that each component contributes uniquely, with TUE and PDI playing critical roles in refining segmentation predictions on the FSD dataset.

\subsubsection{MSD} On the MSD dataset, which contains more challenging small and barely visible tumors, the impact of each component becomes more pronounced. The WSA-only model improves the Dice score by 1.37\% and reduces HD by 0.86 mm compared to the baseline model, highlighting the difficulty of segmenting small structures with basic augmentation alone. Adding BCP slightly enhances performance, increasing the Dice score to 37.23\%, suggesting that while BCP is effective for larger structures, its impact on extremely small tumors is limited.
The introduction of TUE leads to a significant improvement, raising the Dice score to 43.98\% and reducing HD to 21.37 mm, confirming that two-way uncertainty estimation provides crucial guidance for ambiguous tumor regions. IPT, when introduced without TUE, maintains stable performance with a Dice score of 38.22\%, indicating that iterative pseudo-label transition provides additional guidance for small tumors.
The full model, with PDI incorporated, achieves the highest Dice score of 46.61\% and a JA of 37.34\%, demonstrating that prediction disagreement filtering is essential for handling small and uncertain tumor regions. Our results confirm that TUE, IPT, and PDI are key innovations that significantly improve segmentation accuracy, particularly in challenging small tumor scenarios.

We observe performance variations between the FSD and MSD datasets, which we attribute to significant differences in tumor size distribution. Compared to the MSD dataset, the FSD dataset contains larger tumors, as illustrated in Fig.~\ref{fig:overview of tumor data}(a). This benefits BCP-based methods due to their reliance on the ``largest connected component'' strategy. In contrast, tumors in the MSD dataset are generally smaller, allowing our method to achieve more substantial improvements. Nonetheless, results from both datasets confirm the effectiveness of each proposed module.

\subsection{Discussion}
\begin{figure}
  \centering
  \includegraphics[scale=0.7]{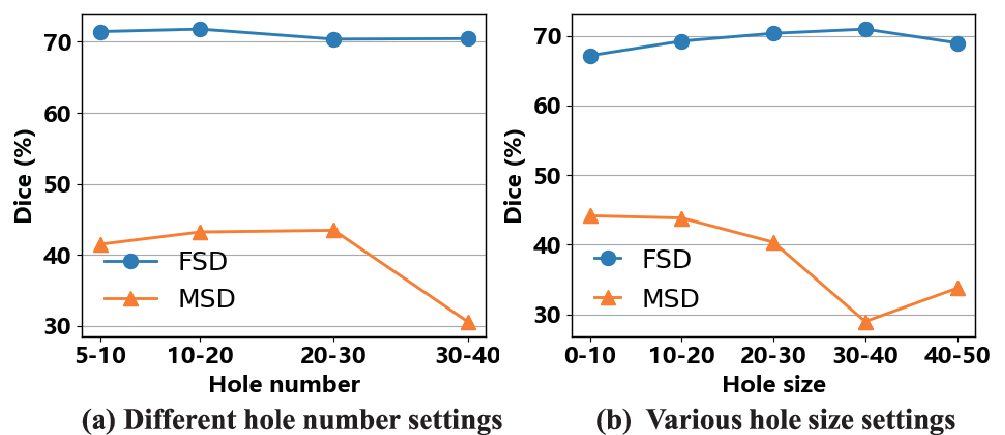}
  \caption{(a) The effectiveness of different hole numbers. (b) The effectiveness of various hole sizes.
  }
  \label{fig:ssl_hole}
\end{figure}

\subsubsection{Effects of hole number and size in $\mathcal{M}$}
We study the impact of varying hole numbers and sizes on the FSD and MSD datasets, as depicted in Fig.~\ref{fig:ssl_hole}. We observe that on the MSD dataset, performance significantly dropped when the number or size of holes exceeded 30. This is simply because an increased number of holes may overlook numerous small tumors on the MSD dataset, thereby diminishing data diversity. In contrast, on the FSD dataset, characterized by relatively larger tumor sizes, the model shows stable performance even with more than 30 holes in a mask. Considering the distinct tumor size distributions and data diversity in each dataset, we randomly generate the hole number for FSD/MSD to range from 5/10 to 40/30, and set the hole size for FSD/MSD from 10/10 to 40/20.

\subsubsection{Effects of threshold $\tau$}
In Table~\ref{table:ablation_study_thres}, we investigate the impact of different threshold values $\tau$ on both datasets. We observe that setting $\tau$ to 0.9 results in optimal performance for both datasets. This is because a higher threshold filters out a substantial number of uncertain voxels, thereby necessitating enhanced discrimination ability from the model. Employing a higher threshold value contributes to incorporating more informative voxels into the augmentation process, effectively encouraging the models to learn more efficiently from unlabeled samples.

\begin{table}
  \centering
  \begin{threeparttable}
    \caption{Segmentation performance on the FSD/MSD dataset with varying values of the threshold parameter $\tau$.}
    \label{table:ablation_study_thres}
    \begin{tabular}{c|ccc|ccc}
      \toprule
      \multirow{2}{*}{Threshold} &  \multicolumn{3}{c|}{FSD} & \multicolumn{3}{c}{MSD} \\ \cline{2-7}
            & Dice~(\%)$\uparrow$ & HD~(mm)$\downarrow$&ASD~(mm)$\downarrow$ &  Dice~(\%)$\uparrow$ & HD~(mm)$\downarrow$&ASD~(mm)$\downarrow$  \\ \hline
      0.5& 70.94 & 7.93&2.40 &	45.42 &	35.57 &18.62\\
      0.6& 71.24 & 8.34&2.46 &43.86 &	16.17 &8.16 \\
      0.7& 68.73 & 10.75 &3.31&	43.67&28.75&18.14 \\
      0.8& 69.87 & 8.71 &2.64 &	43.29 &	24.84&13.42 \\
      0.9& 71.72 &	7.29 &2.05&	46.61& 30.21&17.61 \\
      \bottomrule
    \end{tabular}
  \end{threeparttable}
\end{table}

\begin{table}
  \centering
  \begin{threeparttable}
    \caption{Segmentation performance on the FSD/MSD dataset with different pseudo label strategies.}
    \label{table:ablation_study_moving}
    \begin{tabular}{c|ccc|ccc}
      \toprule
      \multirow{2}{*}{Methods} &  \multicolumn{3}{c|}{FSD} & \multicolumn{3}{c}{MSD} \\ \cline{2-7}
       & Dice~(\%)$\uparrow$ & HD~(mm)$\downarrow$&ASD~(mm)$\downarrow$ &  Dice~(\%)$\uparrow$ & HD~(mm)$\downarrow$&ASD~(mm)$\downarrow$  \\ \hline
      EMA& 68.68 & 9.33 & 2.91 &	44.72& 27.30 & 15.00\\
      VOT& 68.89 & 	9.68 &2.97&	23.53 &	26.37&14.24\\
      IPT& 68.92 &	9.98 &3.07&	46.73  & 32.63 &15.74 \\
      VOT+IPT& 71.72 &	7.29 &2.05&	46.61& 30.21&17.61\\
      \bottomrule
    \end{tabular}
  \end{threeparttable}
\end{table}

\subsubsection{Effects of pseudo labeling strategies}
In Table~\ref{table:ablation_study_moving}, we demonstrate the effectiveness of various pseudo labeling strategies, including EMA, voting (VOT), and IPT. For EMA, the final prediction is updated using the EMA of predictions from teacher and student models. In the VOT approach, we average the predictions from teacher and student models. In practice, we apply VOT during the initial 1/5 of the total  training iterations and then switch to IPT for the remaining 4/5 iterations, denoted as VOT+IPT. The results indicate that employing IPT alone results in substantial improvement on the MSD dataset, while showing a slightly better performance on FSD. Notably, combining VOT with IPT enhances performance on FSD.

\begin{figure}
  \centering
  \includegraphics[scale=0.57]{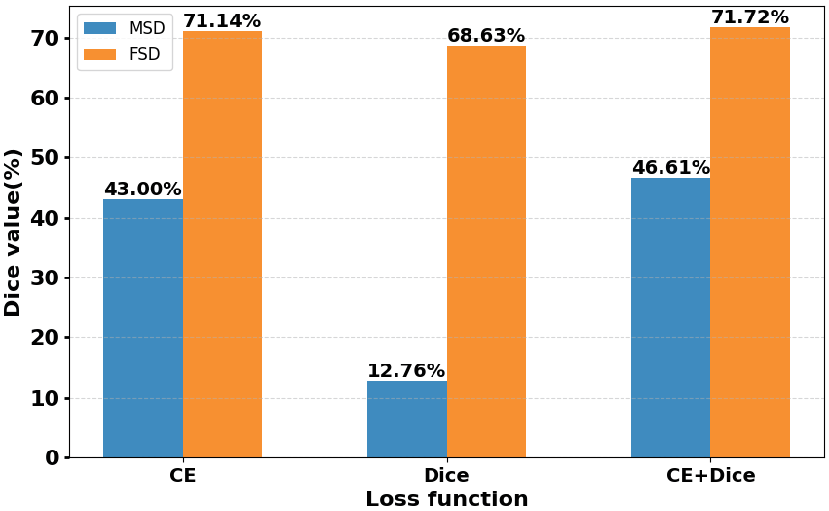}
  \caption{The effectiveness of each loss function.
  }
  \label{fig:ssl_loss}
\end{figure}
\subsubsection{Effects of loss function}
To assess the impact of different loss functions on segmentation performance, we evaluate CE loss, Dice loss, and their combination (CE+Dice) as defined in Eq.~(\ref{eq:loss_func}) on the FSD and MSD datasets, as shown in Fig.~\ref{fig:ssl_loss}. CE loss achieves a Dice score of 43.00\% on MSD and 71.14\% on FSD, demonstrating its effectiveness in optimizing classification-based learning. In contrast, Dice loss yields the lowest performance, with only 12.76\% Dice on MSD and 68.63\% on FSD, likely due to its reliance on overlap-based optimization, which can be unstable for small tumor segmentation. The combination of CE and Dice losses leads to the best performance, reaching 46.61\% Dice on MSD and 71.72\% on FSD, as it benefits from CE's stable optimization and Dice's region-aware segmentation refinement. These results indicate that while CE loss provides strong general supervision and Dice loss focuses on overlap, their combined use effectively balances learning stability and segmentation accuracy, making CE+Dice the most suitable choice for small tumor segmentation in our semi-supervised setting.

\begin{figure}
  \centering
  \includegraphics[scale=0.5]{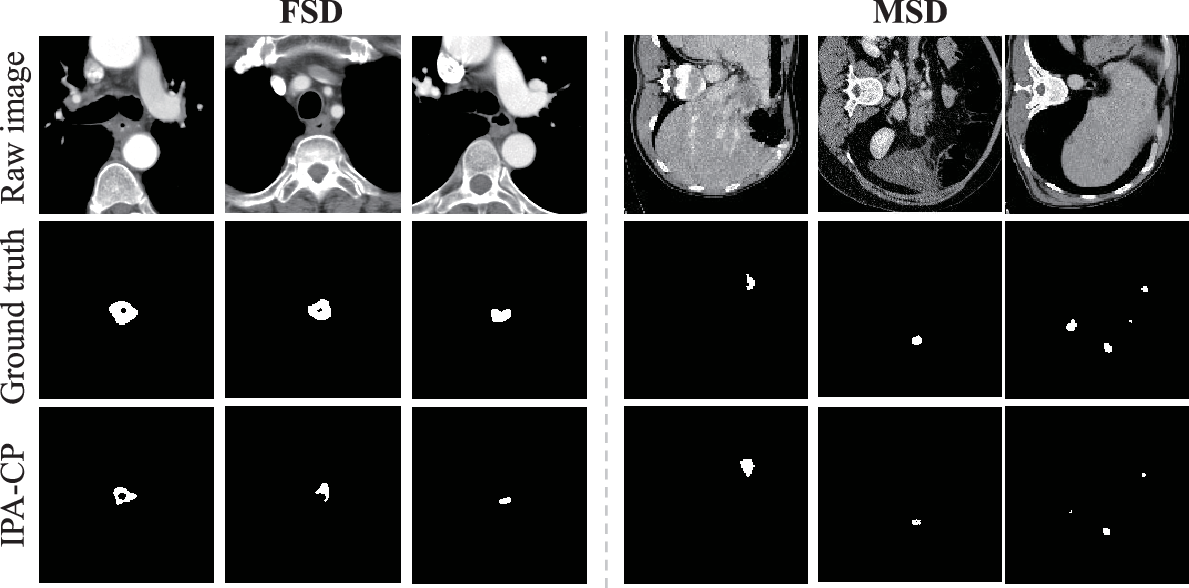}
  \caption{The false prediction cases in extremely small tumors.}
  \label{fig:ssl_fail_tumors}
\end{figure}

\subsubsection{Limitation and future work}
While IPA-CP demonstrates significant advancements in semi-supervised tumor segmentation, several limitations remain. First, although IPA-CP effectively addresses small tumor segmentation, its performance may still be challenged by extremely small or barely visible tumors. For example, while IPA-CP outperforms other methods, false predictions still occur in extremely small tumors compared to the ground truth, as shown in Fig.~\ref{fig:ssl_fail_tumors}. Second, while IPA-CP has been validated on CT scans, its generalizability to other imaging modalities, such as MRI or ultrasound, remains unproven and may require substantial adaptation to maintain efficacy.

To further enhance the effectiveness of IPA-CP, future work will focus on improving its capability in segmenting extremely small or barely visible tumors by refining the pseudo-labeling strategy and incorporating context-aware features that leverage spatial and structural priors. Additionally, to improve the generalizability of IPA-CP across different imaging modalities, we will extend our framework to MRI and ultrasound datasets, investigating modality-specific adaptations and augmentation strategies to ensure robust performance beyond CT imaging.


\section{Conclusion}
In this study, we present an iterative pseudo-labeling based adaptive copy-paste supervision method for semi-supervised tumor segmentation. To augment the diversity and informativeness of unlabeled data, we propose a two-way uncertainty based adaptive augmentation technique. This approach is simple and effective, and can be easily integrated into other models. Additionally, we implement an iterative pseudo-label transition strategy, guiding the predictions towards more robust pseudo labels. Moreover, we incorporate a bidirectional copy-paste mechanism to facilitate model learning from both labeled and unlabeled data. Comprehensive experiments on the in-house esophageal tumor dataset and the public liver tumor dataset demonstrate the efficacy of our method.

\section*{Acknowledgment}
This work was supported by the National Natural Science Foundation of China [Grant No. 62201460, No. 62222311, and No. 62322112], and the Basic Research Programs of Taicang [Grant No. TC2023JC22].

\bibliography{refs}
\end{document}